\documentclass[10pt,twocolumn,letterpaper]{article}

\usepackage{iccv}

\usepackage{tabularx}
\usepackage{booktabs}
\usepackage[ruled,vlined]{algorithm2e}
\usepackage{graphicx}

\usepackage{setspace} 

\usepackage{hyperref}

\title{SAR-TEXT: A Large-Scale SAR Image-Text Dataset Built with SAR-Narrator and A Progressive Learning Strategy for Downstream Tasks}
\author{Yiguo He\thanks{Equal contribution.} \quad Xinjun Cheng\footnotemark[1] \quad Junjie Zhu \quad Chunping Qiu \quad Jun Wang \quad \\ Xichuan Zhang \quad Qiangjuan Huang \quad 
Ke Yang\thanks{Corresponding author. Email: \texttt{yangke13@nudt.edu.cn}} \\
Intelligent Game and Decision Lab, Beijing, China \\
\texttt{yangke13@nudt.edu.cn}
}

\begin{document}
\maketitle
\begin{abstract}
Vision Language Models (VLMs) have achieved remarkable breakthroughs in the field of remote sensing in recent years. Synthetic Aperture Radar (SAR) imagery, with its all-weather capability, is essential in remote sensing, yet the lack of large-scale, high-quality SAR image-text datasets hinders its semantic understanding. In this paper, we construct SAR-TEXT, a large-scale and high-quality dataset consisting of over 130,000 SAR image–text pairs.
To construct the SAR-TEXT dataset, we design the SAR-Narrator framework, which generates textual descriptions for SAR images through a multi-stage strategy. 
To verify the effectiveness of the SAR-TEXT dataset, we conduct experiments on three typical vision-language tasks: image-text retrieval, image captioning, and visual question answering (VQA). Specifically, we construct three representative models on SAR-TEXT: SAR-RS-CLIP, SAR-RS-CoCa, and SAR-GPT. SAR-RS-CLIP achieves notable improvements in retrieval performance, boosting average recall by 12.97\% and 10.0\% on the OSdataset\_512 and HRSID test sets, respectively. In the captioning task, SAR-RS-CoCa achieves significant improvements over the original CoCa models in terms of BLEU-4, SPICE, and CIDEr scores.
In the VQA task, SAR-GPT outperforms baseline and single-stage models on multiple SAR-VQA datasets, demonstrating stronger semantic understanding and reasoning ability, as further confirmed by qualitative results.
\\It is worth noting that, as a flexible captioning tool, SAR-Narrator can be readily adopted by the community to construct larger-scale SAR image–text datasets. All code, pretrained models, and the SAR-Text dataset are publicly available at: \url{https://github.com/YiguoHe/SAR-TEXT}.

\end{abstract}
   
\section{Introduction}
\label{sec:intro}

\begin{figure}[t]
  \centering
   \includegraphics[width=\linewidth]{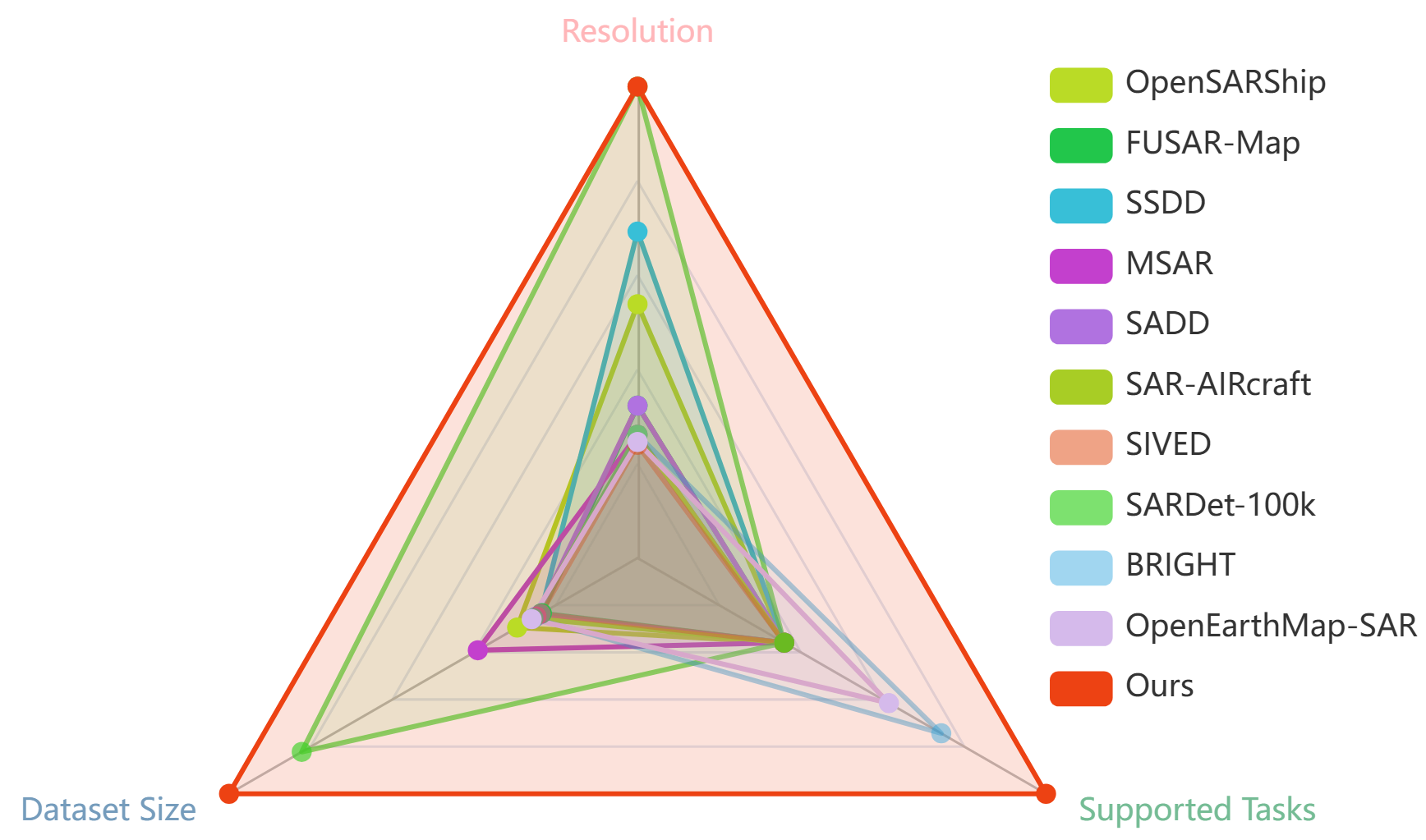}
   \caption{An overall performance comparison between SAR-TEXT and 10 other datasets (OpenSARShip~\cite{huang2017opensarship}, FUSAR-Map~\cite{shi2021object}, SSDD~\cite{zhang2021sar}, MSAR~\cite{xia2022crtranssar}, SADD~\cite{zhang2022sefepnet}, SAR-AIRcraft~\cite{zhirui2023sar}, SIVED~\cite{lin2023sived}, SARDet-100k~\cite{li2024sardet}, BRIGHT~\cite{chen2025bright}, and OpenEarthMap-SAR~\cite{xia2025openearthmap}) across 3 different dimensions at dataset size, resolution, and supported task types. Results demonstrate that SAR-TEXT outperformed existing datasets, showcasing superior and more comprehensive application potential in SAR image interpretation.}
   \label{fig:dataset_comparison}
\end{figure}

\begin{figure*}[t]
  \centering
   \includegraphics[width=\linewidth]{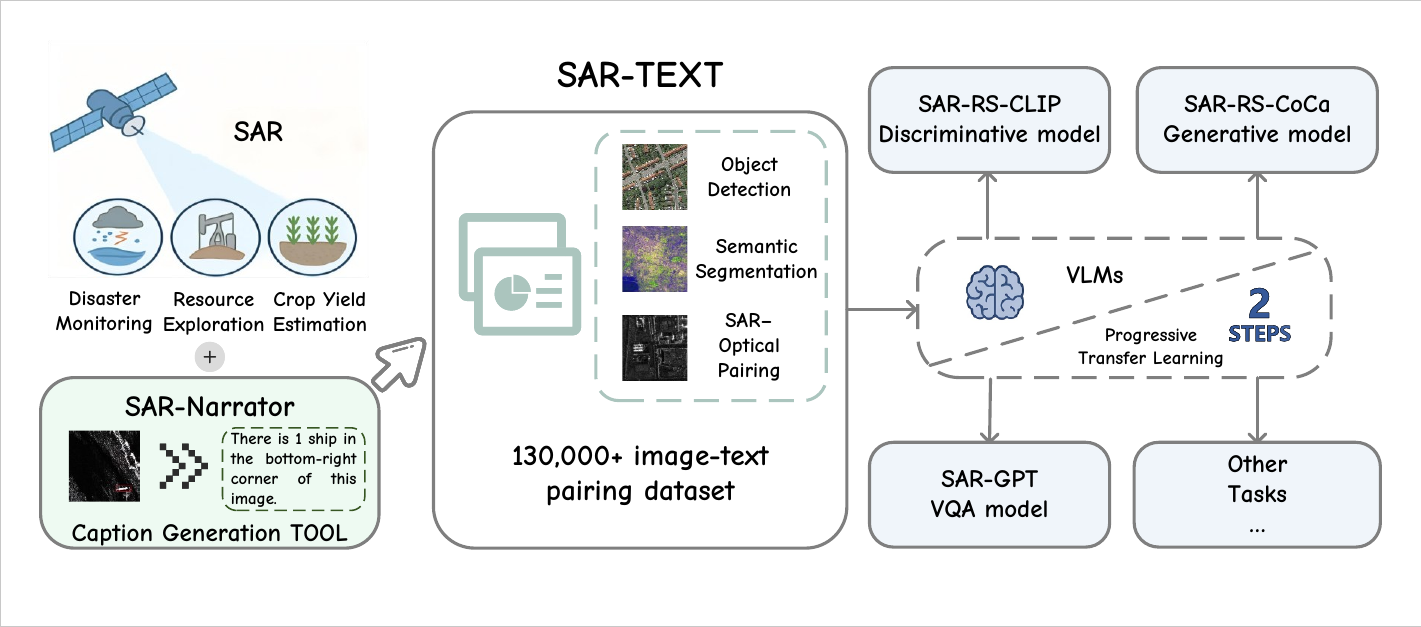}
   \caption{Overview of the SAR-Narrator framework and SAR-TEXT dataset: SAR images are automatically captioned by SAR-Narrator to construct SAR-TEXT, a high-quality dataset containing 130,000+ text-image pairs. Subsequently, two models—SAR-RS-CLIP and SAR-CoCa—are trained using vision-language models (VLMs) and a progressive two-stage fine-tuning strategy to achieve more effective semantic interpretation of SAR images.}
   \label{fig:overview}
\end{figure*}

Synthetic Aperture Radar (SAR) is an active microwave remote sensing technology that is widely recognized for its capability to acquire high-resolution imagery regardless of weather conditions, including cloud cover, precipitation, and darkness~\cite{11}. Unlike the imaging principles of optical images, SAR imaging uses microwave signals of different wavelengths, enabling it to penetrate materials such as vegetation, dry soil, and man-made structures. It is precisely because of SAR's outstanding penetration capabilities, coupled with its excellent adaptability to all weather conditions and harsh environments, that it has a wide range of applications, such as disaster response, resource exploration, and crop yield estimation~\cite{155}. In recent years, the rapid advancement of Large Language Models (LLMs) and Vision-Language Models (VLMs) has made the interpretation of SAR imagery into natural language an increasingly effective approach~\cite{3,8,9}. This method not only facilitates the extraction and communication of critical information from SAR data but also significantly improves its readability and accessibility, particularly for non-experts, thereby enabling a broader audience to understand and utilize complex SAR information.

However, due to the complex imaging mechanisms of SAR and the substantial differences in visual texture compared to optical imagery, semantic interpretation of SAR data has long presented significant challenges. Traditional approaches often rely on expert-driven manual analysis, which is time-consuming, labor-intensive, and insufficient to meet the increasing demand for efficient SAR image interpretation. Consequently, several studies have attempted to employ machine learning techniques to assist in SAR image interpretation, yet their effectiveness remains limited~\cite{156,157}. With the advancement of deep learning techniques~\cite{158,159}, research in remote sensing has made significant progress, particularly through recent developments in vision language foundation models (VLFMs) such as CLIP~\cite{2} and CoCa~\cite{1}, which have achieved remarkable success in optical remote sensing. However, these advancements remain largely confined to optical imagery, as VLFMs continue to struggle with SAR data due to the substantial modality gap and the scarcity of annotated SAR–text pairs~\cite{13}. More notably, the substantial differences between human visual perception and SAR image representations not only increase the complexity of annotation tasks but also necessitate specialized domain knowledge and extensive training for annotators, leading to high annotation costs. Consequently, existing SAR datasets are limited in both scale and diversity. These challenges underscore the importance of constructing large-scale, high-quality SAR image–text annotation datasets and developing efficient algorithms and models specifically designed for SAR imagery.

To address the aforementioned challenges, this paper presents SAR-Narrator, a SAR image caption annotation tool, which is employed to construct a large-scale SAR image–text paired dataset named SAR-TEXT, which contains more than 130,000 SAR images and their corresponding high-quality natural language descriptions, effectively alleviating the problem of scarce annotation data in this field. In addition, a two-stage progressive fine-tuning strategy is proposed to adapt vision–language foundation models, specifically CLIP and CoCa, with the goal of bridging the modality gap between SAR and natural images and enhancing the semantic interpretability of SAR data. Experimental results demonstrate the effectiveness of both the SAR-Narrator tool and the proposed progressive transfer learning approach. Furthermore, additional exploration of downstream VQA tasks has demonstrated the practical applicability of the proposed SAR-TEXT dataset. Details are shown in Figure~\ref{fig:overview}.

Our main contributions are summarized as follows:

\begin{enumerate}
    \item We propose \textbf{SAR-Narrator}, an automated SAR image captioning tool that integrates the generative vision–language foundation model RS-Captioner, A2C and SA2C algorithms, and the in-context learning capabilities of large language models. This framework enables high-quality conversion from structured labels to natural language captions, significantly improving annotation efficiency and semantic diversity.
    \item Based on SAR-Narrator, this study constructs \textbf{SAR-TEXT}, the first large-scale and high-quality SAR image–text paired dataset. It contains more than 130,000 SAR images with precise textual captions, effectively addressing the shortage of multimodal data in SAR interpretation research.
    \item A \textbf{progressive fine-tuning strategy} is introduced for adapting vision language foundation models (VLFMs) to SAR imagery. This leads to the development of SAR-RS-CLIP and SAR-RS-CoCa, the first VLFMs specifically trained for SAR image–text understanding. Experimental results demonstrate that both models substantially outperform baseline methods in SAR image–text retrieval and caption generation tasks.
\end{enumerate}

The remainder of this paper is organized as follows. Section 2 reviews related work. Section 3 provides a detailed description of the SAR-Narrator tool generation method and the SAR-TEXT dataset construction process. Section 4 presents the corresponding experimental results and analysis. Finally, Section 5 draws conclusions.

\section{Related Work}
\label{sec:related work}
While VLMs have achieved notable success in natural image understanding, their application to SAR imagery remains highly challenging. This section reviews related work in two key areas corresponding to the core limitations addressed in this study: (i) SAR image–text datasets, and (ii) the application of VLFMs to remote sensing imagery. 

\subsection{SAR Image-Text Datasets}
Although multimodal learning has gained increasing attention in the field of remote sensing, research on image–text matching for SAR imagery remains highly limited. This limitation is primarily attributed to the inherent challenges of SAR imagery: its semantic content is less intuitive, annotation is costly, and the imaging mechanism introduces unique complications such as speckle noise and structural ambiguity.

Among the limited existing efforts, MMRS-1M~\cite{171} and SSIC~\cite{172} are among the most representative. MMRS-1M is a large-scale multimodal instruction dataset encompassing optical, infrared, and SAR imagery. This dataset integrates 34 remote sensing sub-datasets and converts object detection class annotations into textual instructions through rule-based transformation, thereby enabling unified cross-modal modeling. However, the dataset is not SAR-specific, with SAR–text pairs accounting for only a small fraction of the overall data. Furthermore, the generated captions are predominantly templated, exhibiting limited semantic richness and linguistic diversity.

Zhao et al. made the first attempt to generate natural language descriptions from SAR imagery~\cite{172}. They constructed a private dataset named SSIC and adopted an attention-based encoder–decoder framework to enable automatic caption generation, thereby demonstrating the feasibility of SAR image captioning. However, SSIC is limited to a single object category (ship) and is not publicly available, which restricts its generalizability and reproducibility.

\subsection{VLMs for Remote Sensing}
As previously noted, vision–language models (VLMs) \cite{sun2025mmtp,liu2024text,he2024rethinking,qiu2024few,zhu2024mvp,zhu2023hybrid,zhu2023hcpnet} have shown significant potential for a wide range of applications in the remote sensing domain. Their use spans multiple tasks, including land cover classification and scene understanding, target detection and object recognition in remote sensing imagery, change detection and disaster assessment, semantic retrieval, and information querying.

Recent research on vision–language models for remote sensing has explored several distinct directions. Models such as RemoteCLIP~\cite{remoteclip}, GeoRSCLIP~\cite{georsclip}, and SkyCLIP~\cite{skyscript} focus on foundational model development, enhancing the semantic representation capabilities of remote sensing imagery through large-scale pretraining on image–text pairs. In contrast, RS-CLIP~\cite{RSCLIP} and SenCLIP~\cite{senclip} are designed for specific tasks such as zero-shot classification and land use mapping, leveraging contrastive learning and cross-modal alignment to significantly improve classification performance. ChangeCLIP~\cite{changeclip} introduces vision–language contrastive learning into the domain of change detection, enabling robust identification of changes across diverse scenarios. More recent models, including SkyEyeGPT~\cite{skyeyegpt} and Falcon~\cite{falcon}, represent advances in multi-task integration. SkyEyeGPT incorporates visual features into large language models to support conversational question answering and semantic interpretation of remote sensing imagery. Falcon, meanwhile, aims to unify 14 distinct tasks—including classification, detection, segmentation, and description—within a single framework. Collectively, these efforts reflect the ongoing evolution of remote sensing intelligence from task-specific models toward general-purpose multimodal architectures.

In summary, none of the aforementioned works provide a standardized means of evaluation for SAR image-text matching or retrieval tasks, nor do they provide effective performance validation. This motivates the construction of SAR-TEXT, a high-quality, large-scale, and diverse dataset of SAR image–caption pairs designed to support both retrieval and captioning tasks. Meanwhile, a two-stage progressive transfer learning strategy is proposed for remote sensing VLFMs, which leverages optical remote sensing imagery as a 'springboard' to gradually facilitate knowledge transfer. This approach helps mitigate the modality gap between optical and SAR images, thereby enabling VLFMs to more effectively perform semantic interpretation tasks on SAR imagery.

\begin{figure*}[t]
  \centering
   \includegraphics[width=\linewidth]{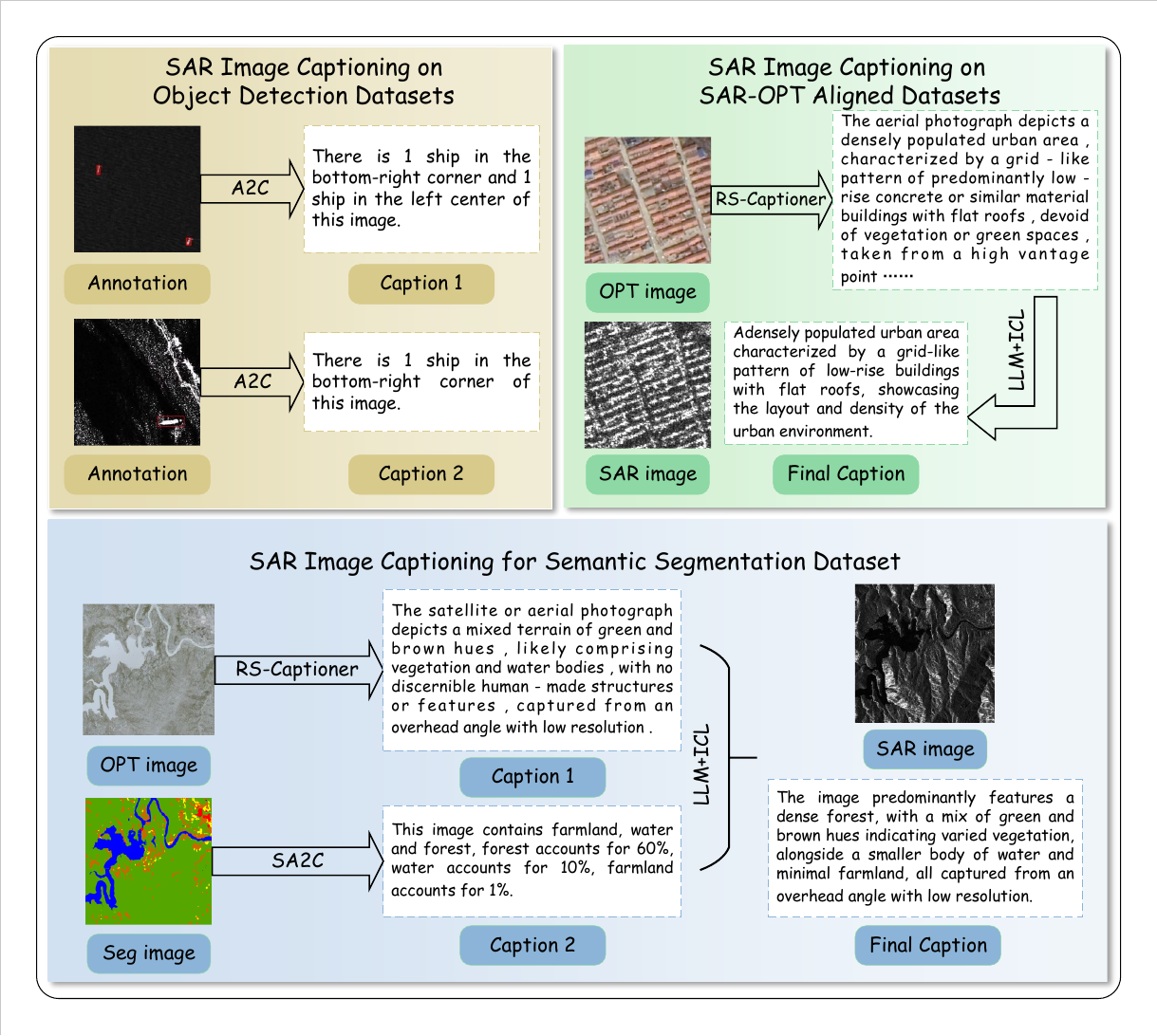}
   \caption{SAR-TEXT dataset construction method.}
   \label{fig:dataset}
\end{figure*}

\section{Methodology}

To address the challenges of automatic semantic interpretation for SAR imagery, this paper proposes an automated caption generation framework named SAR-Narrator. The overall architecture of the SAR-Narrator method is illustrated in Figure \ref{fig:dataset}. Constructed using the SAR-Narrator framework, the SAR-TEXT dataset provides a robust foundation for intelligent interpretation and multimodal applications of SAR imagery. It is also worth noting that a progressive transfer learning strategy is introduced, enabling VLFMs to achieve strong performance on SAR imagery even under limited annotated data conditions.

\subsection{SAR-Narrator Framework}
Given the high level of abstraction in semantic expression of SAR images and the high cost of manual annotation, this paper designs and implements an automated subtitle generation framework named SAR-Narrator. The framework integrates multiple strategies, employing the A2C algorithm (object detection), SA2C algorithm (semantic segmentation), and a context-learning rewriting mechanism based on large language models (SAR–optical pairing) for different types of datasets, to achieve high-quality, low-cost natural language description generation. During the construction of the SAR-TEXT dataset, this paper selected DeepSeek-V3~\cite{4} as the large language model for caption rewriting, effectively enhancing the naturalness and modality adaptability of the descriptions. The ultimately constructed SAR-TEXT provides critical data support for the intelligent interpretation of SAR images and significantly reduces the cost of manual annotation.

\subsubsection{A2C Algorithm}
We propose an Annotation-to-Caption (A2C) algorithm designed to transform object detection annotations into descriptive captions. The core steps of the A2C algorithm include parsing the categories, quantities, and spatial locations of objects within SAR imagery, followed by automatically generating structured textual descriptions based on detection boxes (e.g., “There are 3 ships in the top-left of this image.”). The A2C algorithm is particularly suitable for target detection datasets, such as MSAR-1.0~\cite{166} and HRSID~\cite{160}, enriching SAR target recognition tasks with detailed textual annotations. The pseudocode of the A2C algorithm is provided in Algorithm \ref{alg:a2c}.
\begin{algorithm}[]\small
\caption{Annotation to Caption Algorithm (A2C)}
\label{alg:a2c}
\KwIn{List of detected objects with class labels: \texttt{objects}}
\KwOut{Natural language caption: \texttt{caption}}

\BlankLine
\textbf{Initialize:} \\
\Indp
\texttt{classes} $\leftarrow$ \texttt{get\_unique\_classes(objects)} \\
\texttt{class\_counts} $\leftarrow$ \texttt{count\_objects\_by\_class(objects)} \\
\texttt{caption} $\leftarrow$ \texttt{""} \\
\Indm

\BlankLine
\If{\texttt{len(classes) == 1}}{
    \texttt{class} $\leftarrow$ \texttt{classes[0]} \\
    \texttt{count} $\leftarrow$ \texttt{class\_counts[class]} \\
    \eIf{\texttt{count == 1}}{
        \texttt{caption} $\leftarrow$ ``There is 1'' + \texttt{get\_class\_name(class)} + `` in this image.'' 
    }{
        \eIf{\texttt{count == 2}}{
            \texttt{caption} $\leftarrow$ ``There are 2'' + \texttt{get\_class\_name(class)} + `` in this image.''
        }{
            \eIf{\texttt{count <= 10}}{
                \texttt{caption} $\leftarrow$ ``There are '' + \texttt{count} + `` '' + \texttt{get\_class\_name(class)} + `` in this image.''
            }{
                \texttt{caption} $\leftarrow$ ``There are more than ten '' + \texttt{get\_class\_name(class)} + `` in this image.''
            }
        }
    }
}
\Else{
    \ForEach{\texttt{class in classes}}{
        \texttt{count} $\leftarrow$ \texttt{class\_counts[class]} \\
        \eIf{\texttt{count == 1}}{
            \texttt{caption} += ``There is 1'' + \texttt{get\_class\_name(class)} + `` in this image. ''
        }{
            \eIf{\texttt{count <= 10}}{
                \texttt{caption} += ``There are '' + \texttt{count} + `` '' + \texttt{get\_class\_name(class)} + `` in this image. ''
            }{
                \texttt{caption} += ``There are more than ten '' + \texttt{get\_class\_name(class)} + `` in this image. ''
            }
        }
    }
}

\BlankLine
\textbf{Return} \texttt{caption}

\end{algorithm}

\subsubsection{SA2C Algorithm}
Semantic segmentation datasets (e.g., WHU-OPT-SAR~\cite{173}) provide pixel-level classification annotations, serving as valuable resources for SAR image interpretation tasks. To effectively utilize this information, we propose a Segmentation Annotation-to-Caption (SA2C) algorithm, aiming to generate more precise textual descriptions for SAR imagery. The SA2C algorithm initially generates descriptive captions based on the proportional information of semantic classes in segmentation annotations (e.g., “This image contains farmland, water, and forest, with forest accounting for 81\%, water 1\%, and farmland 1\%.”). The pseudocode of the SA2C algorithm is detailed in Algorithm \ref{alg:sa2c}.

\begin{algorithm}\small
\caption{Segmentation Annotation to Caption Algorithm (SA2C)}
\label{alg:sa2c}
\KwIn{\\
\quad Image array \texttt{I}, \\
\quad RGB-to-category mapping: \texttt{rgb\_to\_category},\\
\quad Threshold proportion: \texttt{threshold\_proportion} (default = 1\%)
}
\KwOut{Caption describing the image}
\BlankLine
\textbf{Step 1: Initialization} \\
\Indp
\texttt{total\_pixels} $\leftarrow$ \texttt{image\_array.shape[0]} $\times$ \texttt{image\_array.shape[1]} \\
\texttt{category\_counts} $\leftarrow$ \{category: 0 for category in \texttt{rgb\_to\_category.values()}\} \\
\Indm

\BlankLine
\textbf{Step 2: Count pixels per category} \\
\Indp
\ForEach{(rgb, category) in \texttt{rgb\_to\_category}}{
    \texttt{mask} $\leftarrow$ pixels in \texttt{image\_array} matching \texttt{rgb} \\
    \texttt{category\_counts[category]} += \texttt{sum(mask)}
}
\Indm

\BlankLine
\textbf{Step 3: Compute category proportions} \\
\Indp
\texttt{proportions} $\leftarrow$ \{category: (\texttt{category\_counts[category]} / \texttt{total\_pixels}) $\times$ 100 \\
\hspace*{5em} for category in \texttt{category\_counts}\}
\Indm

\BlankLine
\textbf{Step 4: Filter categories by threshold} \\
\Indp
\texttt{filtered\_proportions} $\leftarrow$ \{category: proportion for (category, proportion) in \texttt{proportions} \\
\hspace*{10em} if \texttt{proportion} $\geq$ \texttt{threshold\_proportion}\}
\Indm

\BlankLine
\textbf{Step 5: Generate caption} \\
\If{\texttt{filtered\_proportions} is not empty}{
    \texttt{included\_categories} $\leftarrow$ categories in \texttt{filtered\_proportions} \\
    \texttt{included\_sentence} $\leftarrow$ join categories into sentence \\
    \texttt{sorted\_proportions} $\leftarrow$ sort \texttt{filtered\_proportions} descending \\
    \texttt{sorted\_sentence} $\leftarrow$ generate sentence from \texttt{sorted\_proportions} \\
    \texttt{caption} $\leftarrow$ ``This image contains '' + \texttt{included\_sentence} + ``. '' + \texttt{sorted\_sentence}
}
\Else{
    \texttt{caption} $\leftarrow$ ``No significant categories found.''
}

\BlankLine
\Return \texttt{caption}

\end{algorithm}

To enhance the accuracy and naturalness of SAR image captions, we further utilize the RS-Captioner model to generate captions from paired optical imagery, extracting supplementary semantic information to support subsequent caption integration. Based on this, the DeepSeek-V3 large language model is introduced to semantically fuse the initial SAR captions with these additional optical-based descriptions, reconstructing them into coherent and comprehensive final captions.

To ensure the quality and consistency of the generated captions, we establish a systematic set of text integration principles, which are provided to the large language model (LLM) as guiding prompts. The detailed prompt design is presented below:

\textbf{Rewrite the existing description to integrate object category proportions and their visual interpretations by following these principles: }

[1] Avoid specific numbers: Refrain from using specific numerical percentages. Instead, use qualitative terms such as “dominates,” “covers a significant portion,” or “forms the majority” to describe proportions. 

[2] Use Specific Terminology: If the description provides more specific terms (e.g., “river” or “lake” instead of the general “water”), prioritize these specific terms and exclude the more general terms from the first description. 

[3] Emphasize Dominant Features: Prioritize the most visually dominant features in the image, and mention the less prominent elements afterward. If multiple categories have similar proportions, mention them in descending order of visual significance. 

[4] Ensure Clarity and Fluidity: The final sentence should be concise, clear, and read like a natural image caption. It should summarize the visual content effectively while maintaining fluency. Avoid redundancy. 

[5] Describe All Major Categories: If the description lists multiple categories, ensure the final sentence reflects all significant elements, even if their proportions are small, but ensure brevity.

During the caption fusion process, we combine and optimize information from multiple sources. For instance, original caption A states: “This image contains farmland, village, and water. Water accounts for 88\%, farmland accounts for 3\%, and village accounts for 1\%.” Original caption B describes: “The image presents an aerial view of a field, captured from a high angle. The field is divided into sections by a network of roads or pathways, creating a grid-like pattern.” Through the fusion process, redundant information is removed while essential scene details are retained, resulting in the optimized caption: “The image showcases a vast water body dominating the scene, with a field divided into sections by a network of roads forming a grid-like pattern.” This fusion approach ensures that captions accurately reflect the main content of SAR images, enhancing both fluency and readability.

\subsubsection{Rule-Guided LLM-ICL Rewriting for SAR Captioning}
For the SAR-optical image pairing dataset, we consider that while directly annotating SAR images presents significant challenges, generating high-quality descriptive text for the corresponding optical images can serve as an effective indirect semantic annotation method, as the descriptive text generated for the optical images can effectively approximate the semantic content of the corresponding SAR images. However, due to substantial differences between SAR and optical imagery in imaging modality, texture characteristics, and semantic representation, directly transferring captions from optical images may introduce semantic bias and descriptive inaccuracies.

To address these challenges, we propose a caption rewriting mechanism leveraging the semantic understanding capabilities of large language models. Initially, the CoCa model is fine-tuned on the HQRS-IT-210K dataset to construct RS-Captioner~\cite{rethinkingrsclip, he2025enhancingremotesensingvisionlanguage, qiu2024few}, a caption generator specifically designed for remote sensing images, producing preliminary textual descriptions of optical imagery. The high-quality captions generated by RS-Captioner provide a reliable textual foundation for subsequent SAR image caption rewriting based on large language models. Subsequently, a large language model, guided by carefully designed prompts and manually annotated examples through an In-Context Learning (ICL) strategy~\cite{170}, semantically adjusts and optimizes these initial captions to generate refined natural language descriptions that more accurately reflect the semantic characteristics of SAR imagery.

This strategy effectively integrates the interpretability of optical imagery with the language comprehension and transfer capabilities of large language models (LLMs), enabling high-quality and cost-efficient caption annotation of SAR imagery. The resulting SAR-TEXT dataset exhibits strong semantic consistency and broad applicability, providing essential data support for subsequent SAR-based multimodal model training.

To enhance the modal adaptability of subtitles, this paper further introduces the context learning method of large language models to perform targeted rewriting of optical image subtitles, making them more consistent with the visual semantic features of SAR images. In terms of rewriting strategies, this paper systematically analyzes the differences between SAR images and optical images in imaging mechanisms and visual representation, deriving a set of general principles applicable to the rewriting task. These principles are then converted into guided prompts to support the LLM in achieving precise semantic transfer and language generation under contextual conditions. The specific prompt design for rewriting is as follows.

\textbf{Rewrite the existing description to suit Radar images by following these principles: }

[1] Remove color descriptions, such as gray, black, shades and white.

[2] Remove speculative or tentative descriptions, such as “possibly buildings or storage facilities.” 

[3] Preserve primary visual objects but omit descriptions of trees. 

[4] Remove irrelevant details unrelated to visual objects, such as references to camera properties or imaging conditions.

To improve the LLM's rewriting ability, this paper first manually annotated approximately 50 examples of rewriting optical remote sensing images into SAR images. During the model rewriting process, optimization was performed by combining rewriting principles with N randomly selected examples. For the selection of N, this paper balanced the adequacy of learning examples with the LLM's contextual reading ability, ultimately setting N=3 in the experiments.

For example, during the specific rewriting process, when the input is “The black and white aerial photograph depicts a landscape divided into two distinct sections by a diagonal line, with a large, rectangular farm or agricultural area on the left and a densely vegetated area on the right,” the model outputs the adjusted description as “A landscape divided by a diagonal line, with a large farm on the left and a densely vegetated area on the right.” This adjustment ensures that SAR captions align with the characteristics of SAR images while preserving as much semantic information from optical images as possible, thereby enhancing the quality and applicability of the SAR-TEXT dataset.

\subsection{SAR-TEXT Dataset Construction}

\begin{table*}[!htbp]
\centering
\caption{Summary of SAR Datasets}
\label{tab:sar_datasets}
\renewcommand{\arraystretch}{1.2}
\begin{tabular}{p{2.5cm} c p{2.8cm} p{8.5cm}}
\toprule
\textbf{Dataset} & \textbf{Counts} & \textbf{Class} & \textbf{Description} \\
\midrule
MSAR-1.0 & 28449 & Aircraft, Oil Tank, Bridge, Ship & Uses Gaofen-1 and Gaofen-3 data, with HH, HV, VH, VV polarizations. Includes scenes like airports, ports, islands, with targets such as aircraft, oil tanks, bridges, and ships. \\
\midrule
SAR-ship & 43819 & Ship & The dataset is composed of 102 images from China’s Gaofen-3 satellite and 108 images from Sentinel-1, containing 43,819 ship chips, each with a resolution of 256×256 pixels, with both range and azimuth dimensions of 256 pixels. \\
\midrule
OSdataset & 11245 & - & The dataset includes 2673 pairs of 512×512 pixel registered images and 10692 pairs of 256×256 pixel images. The SAR-TEXT training dataset uses only the annotated training and validation images. Optical images are from Google Earth, and SAR images are from China’s Gaofen-3 satellite with 1-meter resolution in C-band full polarization. \\
\midrule
HRSID\_JPG & 3642 & Ship & The images are acquired from China’s Gaofen-3 satellite, containing 5,604 high-resolution SAR images and 16,951 annotated ship instances, with resolutions of 0.5 meters, 1 meter, and 3 meters. \\
\midrule
QXSLAB\_SAROPT & 20000 & - & The SAR images are from China’s Gaofen-3 satellite, and the optical images are from Google Earth. The dataset includes 20,000 pairs of corresponding image tiles collected from three port cities: Jeddah, Shanghai, and Qingdao. \\
\midrule
SEN12 & 18094 & - & The original dataset is collected from multiple satellites, covering global regions and seasonal variations, containing 282,384 pairs of corresponding image tiles. After filtering, 18,094 pairs of samples were selected for annotation. \\
\midrule
optical\_sar & 8575 & - & The original dataset contains 10,000 pairs of SAR and optical images for regional analysis. After processing and selection, 8,575 images were used for annotation. \\
\midrule
whu-sar-opt & 1600 & Farmland, City, Village, Water, Forest, Road & The optical images in the dataset are from China’s Gaofen-1 satellite, and the SAR images are from China’s Gaofen-3 satellite. The original data consists of 100 pairs of optical and SAR images, each approximately 5556×3704 pixels, covering an area of about 50,000 square kilometers in Hubei Province. After slicing the images using a sliding window, 1,600 pairs of samples were obtained. \\
\midrule
SSDD & 1160 & Ship & The original dataset contains 5,604 cropped SAR images, from which 1,160 images were selected for annotation after filtering. \\
\midrule
\textbf{Total} & \textbf{136584} & - & Total sample count. \\
\bottomrule
\end{tabular}
\end{table*}

\subsubsection{Data Preparation}
To construct a high-quality SAR image–text dataset, this paper integrates multiple publicly available SAR remote sensing datasets spanning diverse task types, including target detection, semantic segmentation, and SAR–optical image pairing. This integration ensures diversity in content, semantics, and modalities. The primary data sources can be categorized into the following three groups:

\textbf{Object Detection Datasets:} These include HRSID~\cite{160}, SAR-Ship~\cite{161}, SSDD~\cite{162}, and MSAR-1.0~\cite{166}, which primarily provide object localization and category annotations. They are well-suited for small object recognition and object-level semantic interpretation tasks.

\textbf{Semantic Segmentation Dataset:} This category includes the WHU-OPT-SAR dataset~\cite{173}, which provides pixel-level segmentation annotations for SAR–optical image pairs. It is well-suited for scene understanding and multimodal collaborative analysis tasks.

\textbf{SAR–Optical Pairing Datasets:} This group includes the OS dataset~\cite{174}, QXS-SAROPT~\cite{175}, SEN1-2~\cite{176}, and SARptical~\cite{177}, all of which provide co-registered SAR and optical image pairs. These datasets support cross-modal modeling and the construction of image–text correspondences.

These datasets provide rich structured annotations and semantically diverse imagery, laying a solid foundation for the subsequent generation of high-quality image-text paired samples. Detailed information about these datasets is presented in Table  \ref{tab:sar_datasets}.

After acquiring the images, the dataset first undergoes data cleaning to remove corrupted samples and those lacking corresponding annotations. The perceptual hash (p-hash) algorithm~\cite{5} is then applied for strict deduplication to ensure that no duplicate images exist in the dataset. Additionally, higher p-hash similarity thresholds are selectively applied to individual sub-datasets to eliminate highly similar samples, thereby enhancing the overall diversity and variability of the data.

\subsubsection{SAR-TEXT Dataset Description}
\begin{figure}[h]
  \centering
   \includegraphics[width=\linewidth]{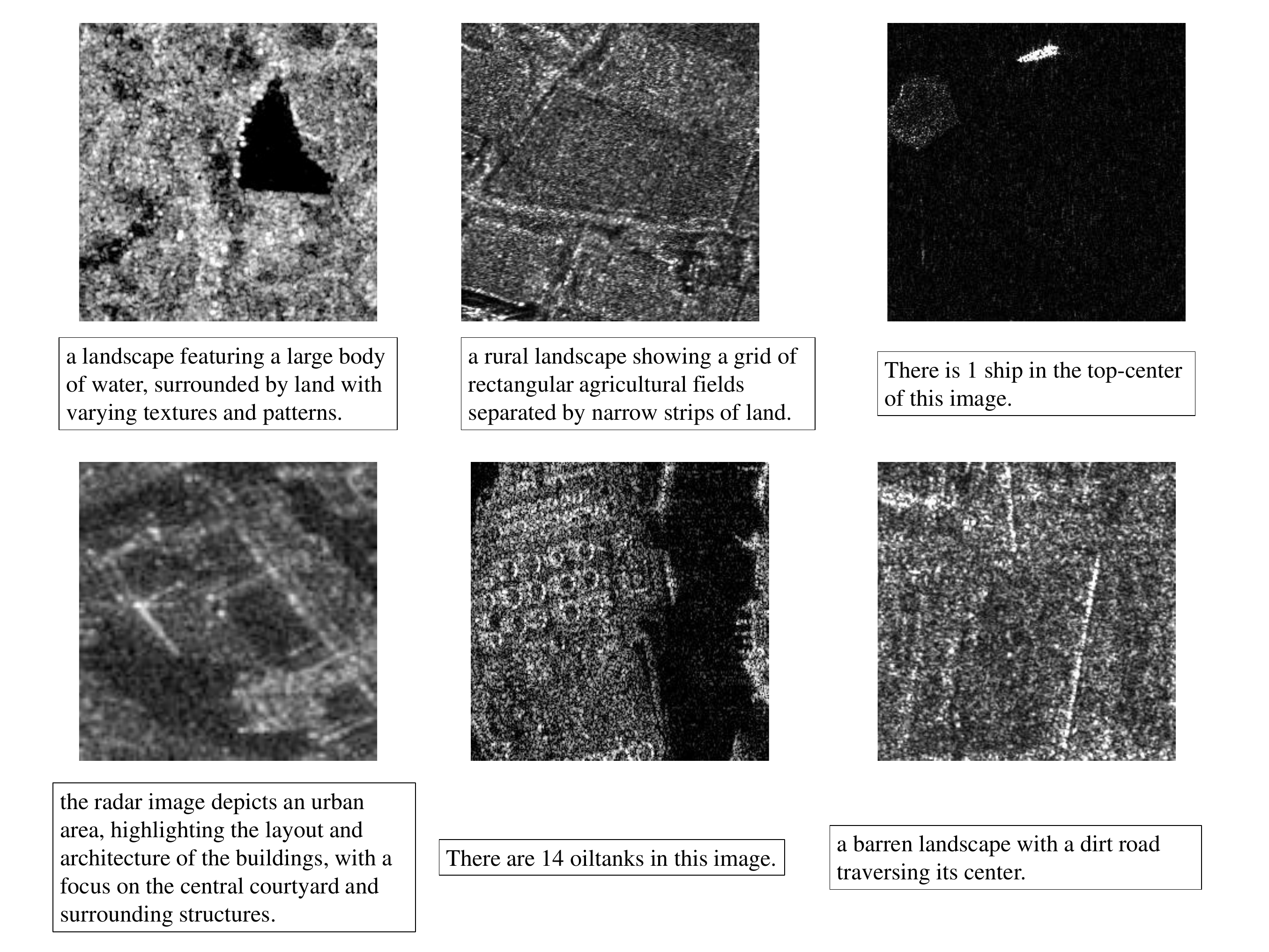}
   \caption{SAR-TEXT dataset sample example.}
   \label{fig:sample example}
\end{figure}

Through the above methods, the SAR-TEXT dataset is automatically annotated, significantly reducing labor costs while ensuring the accuracy and consistency of the caption content. The final SAR-TEXT dataset contains 136,584 SAR images paired with high-quality textual descriptions, sourced from multiple publicly available SAR datasets and exhibiting strong semantic diversity and scene coverage. Each caption has an average length of 16.3 English words with high information density, effectively capturing the semantic characteristics of the corresponding image. This dataset provides essential support for SAR image interpretation and downstream applications, laying a foundation for the integration of multimodal models in remote sensing. Representative examples of image–text pairs are illustrated in Figure \ref{fig:sample example}, visually demonstrating the high quality of the dataset.

\begin{figure}[h]
  \centering
   \includegraphics[width=0.8\linewidth]{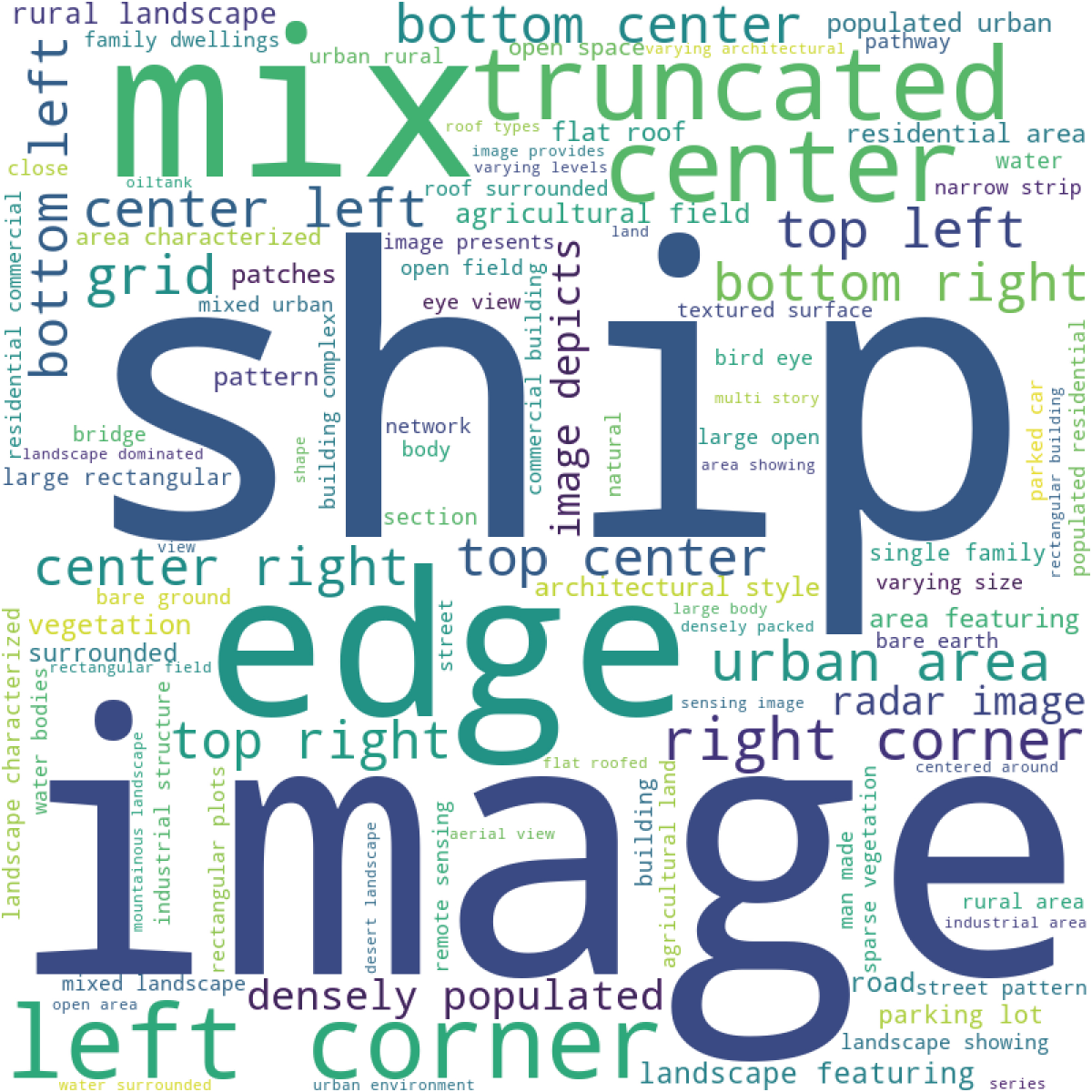}
   \caption{SAR-TEXT dataset word cloud.}
   \label{fig:wordcloud}
\end{figure}

\begin{figure}[h]
  \centering
   \includegraphics[width=\linewidth]{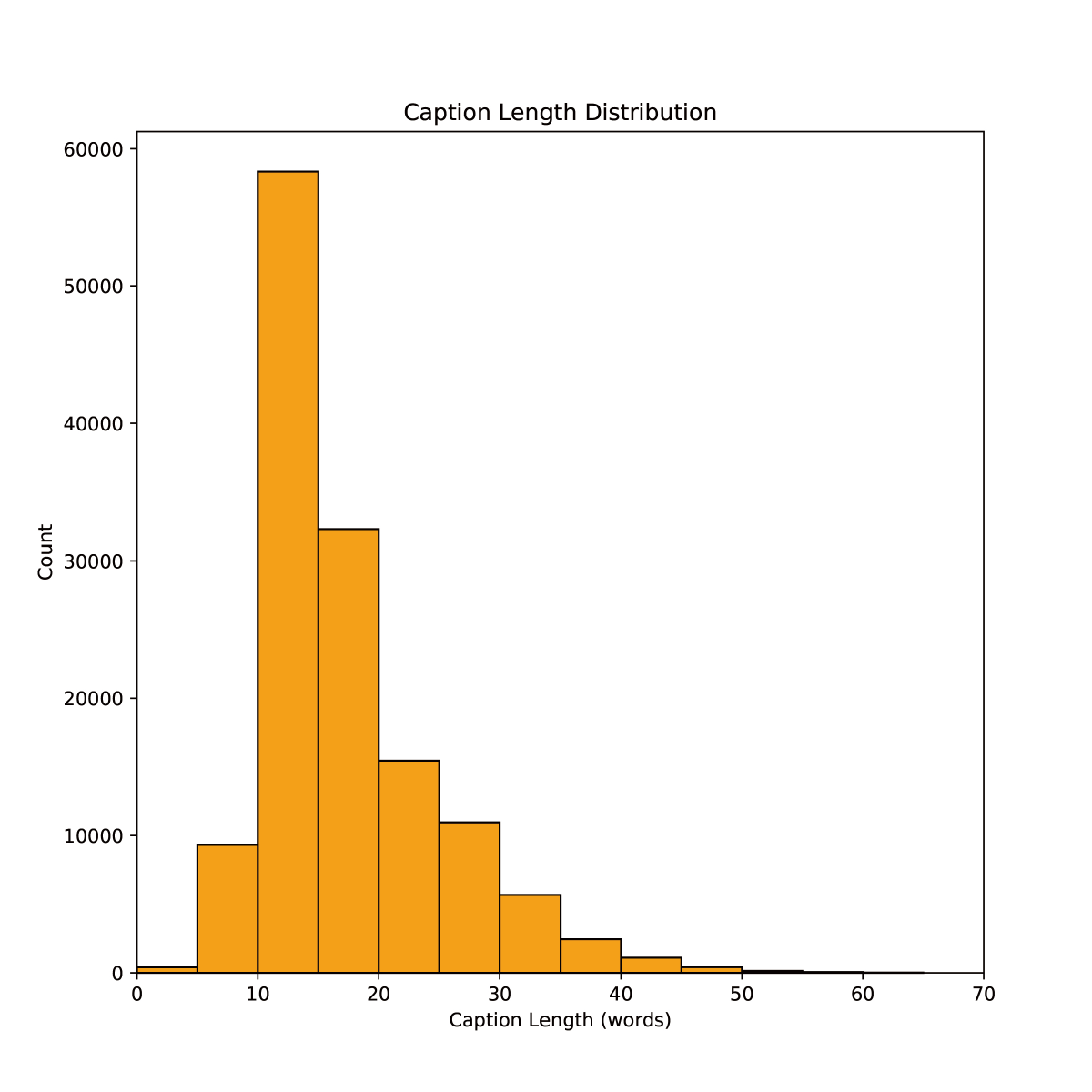}
   \caption{SAR-TEXT dataset caption length distribution.}
   \label{fig:length distribution}
\end{figure}

As illustrated in Figures \ref{fig:wordcloud} and \ref{fig:length distribution}, the dataset exhibits substantial richness in both vocabulary distribution and caption length, highlighting its advantages in semantic diversity and fine-grained descriptive capability. The word cloud analysis shows that the term 'ship' appears with the highest frequency, indicating that ship detection is a central theme of the dataset. This prominence stems from the inclusion of multiple representative SAR ship detection datasets, such as HRSID, SAR-Ship, SSDD, and MSAR-1.0. Moreover, the word cloud features numerous spatial descriptors—such as 'left corner' and 'right corner'—demonstrating that the captions convey not only object categories but also detailed spatial positioning. The presence of scene-related terms like 'urban area' and 'grid' further suggests that the dataset spans a variety of geographic environments, thereby reinforcing its fine-grained semantic expressiveness.

Analysis of caption length distribution reveals that the majority of captions exceed 10 words, with over half containing 20 words or more, indicating a predominance of longer captions. This reflects the dataset's ability to provide rich scene-level descriptions that go well beyond the expressive scope of traditional short labels. Such detailed annotations facilitate the learning of complex scene semantics by vision–language models, thereby enhancing their interpretability and generalization capability.

Overall, the dataset demonstrates strong category diversity, semantic richness, and fine-grained descriptive capability, making it well-suited for a variety of tasks, including image captioning, object detection, semantic segmentation, and image retrieval. It offers a high-quality data foundation for advancing automated description research in SAR image–text understanding.

\subsection{Progressive Transfer Learning Strategy}

In the field of remote sensing, particularly for SAR image recognition tasks, transfer learning techniques are widely employed to address data scarcity caused by the substantial modality differences between optical and SAR imagery, as well as the lack of large-scale, high-quality annotated datasets. 

Inspired by the concept of transitive transfer, we propose a progressive transfer learning strategy for training vision–language foundation models (VLFMs) on SAR imagery. This approach involves two-stage fine-tuning of the CLIP and CoCa models, enabling them to effectively transfer knowledge from the natural image domain to SAR image recognition and interpretation tasks. The two stages are as follows:
\begin{itemize}
    \item \textbf{Stage 1:} Pretrain CLIP and CoCa on HQRS-IT-210K (optical RS data);
    \item \textbf{Stage 2:} Fine-tune the pretrained models on SAR-TEXT.
\end{itemize}

For training the SAR-RS-CLIP model, the CLIP backbone is first continual pretraining on the HQRS-IT-210K dataset to adapt it to optical remote sensing image–text matching tasks. It is then further fine-tuned on the SAR-TEXT dataset to complete the transfer to SAR image–text matching. The SAR-RS-CoCa model follows the same two-stage training procedure, with the CoCa backbone also continual pretraining on HQRS-IT-210K and subsequently fine-tuned on SAR-TEXT. Ablation experiments confirm the effectiveness of this progressive transfer strategy.

\section{Experiments and Analysis}
This section primarily introduces the experimental design and results analysis of this paper in the training and evaluation of visual language foundation models for SAR images. First, we describe the experimental configurations, including model architecture, training parameters, and hardware environment, and explain the evaluation metrics used in image caption generation and cross-modal image-text retrieval tasks. Subsequently, the performance on optical and SAR images is systematically demonstrated and analyzed through quantitative evaluation, qualitative examples, and feature embedding visualization, providing a comprehensive assessment of the model's semantic understanding capabilities. Finally, we validate the effectiveness and necessity of the progressive transfer learning strategy proposed in this paper through ablation experiments.

\subsection{Experimental Setup}
To establish a vision-language foundation for SAR imagery, two widely recognized and promising VLFMs—CLIP and CoCa—are selected as the primary experimental models. For the CLIP model, the CLIP ViT-L-14 architecture is adopted with a full fine-tuning strategy. Training is conducted on a single RTX 4090 GPU (24 GB VRAM) without employing data augmentation or hyperparameter tuning to maintain experimental simplicity. A cosine learning rate scheduler, mixed-precision training (AMP mode), and the AdamW optimizer with a learning rate of $1 \times 10^{-6}$ and a batch size of 32 are utilized. For the CoCa model, the CoCa ViT-L-14 architecture is employed, and optimization is performed using the AdamW optimizer. The contrastive loss weight is set to 1, and the caption generation loss weight is set to 2. The learning rate is set to $2 \times 10^{-5}$, with a batch size of 32. Training is also conducted on a single RTX 4090 GPU (24 GB VRAM). All experiments utilize the OpenCLIP open-source codebase to ensure effective comparison and evaluation across various tasks and models.

\subsection{Evaluation Metrics}
To comprehensively evaluate model performance, this paper selected two test datasets: HRSID-test and OSdataset\_512-test, which represent different application scenarios of SAR images in various tasks. The HRSID-test dataset is derived from the HRSID dataset and contains high-resolution SAR ship images. These images cover a variety of marine environments with high background complexity, effectively testing the model's ability to identify ship features. The OSdataset\_512-test is derived from the OS dataset, covering diverse scenarios such as urban areas and ports, and is specifically designed for image-text retrieval tasks. It tests the model's cross-modal matching capabilities in complex backgrounds. This paper annotated both test sets in the same manner as the SAR-TEXT dataset, serving as a benchmark for evaluation. The diversity of the test sets provides a reliable basis for a comprehensive assessment of model performance.

For the CLIP-series models, evaluation metrics focus on cross-modal retrieval performance, specifically Image-to-Text Recall@K (i2t-R@K), Text-to-Image Recall@K (t2i-R@K), and Mean Recall. The i2t-R@K metric quantifies the proportion of correct textual matches among the top K texts retrieved for a given image, while t2i-R@K assesses the accuracy of the top K images retrieved for a given textual query. Values of K are set to 1, 5, and 10 to comprehensively evaluate retrieval performance at different scales. Mean Recall, calculated as the average of all Recall@K metrics, provides an overall assessment of model effectiveness. These metrics collectively quantify the precision and robustness of CLIP models in SAR image retrieval tasks, particularly emphasizing their cross-modal alignment capability under complex scenarios.

For the CoCa-series models, the evaluation focuses on their ability to generate captions for SAR images. Multiple standard metrics are employed to assess caption quality across various linguistic and semantic dimensions. BLEU (Bilingual Evaluation Understudy) evaluates text fluency and syntactic accuracy by measuring n-gram overlap between generated captions and reference texts. METEOR (Metric for Evaluation of Translation with Explicit Ordering) accounts for morphological variations, semantic similarity, and word order, offering a more comprehensive evaluation. ROUGE-L (Recall-Oriented Understudy for Gisting Evaluation) measures similarity based on the longest common subsequence (LCS), making it suitable for assessing content completeness in longer texts. CIDEr (Consensus-based Image Description Evaluation) quantifies consensus between the generated caption and multiple reference captions by emphasizing the weighting of salient keywords. SPICE (Semantic Propositional Image Caption Evaluation) evaluates scene understanding and semantic logic by constructing and comparing scene graphs. Together, these metrics provide a multidimensional evaluation of CoCa's performance in SAR image captioning, highlighting its strengths in semantic richness and fine-grained detail capture.

\subsection{Experimental Results}
\begin{table*}[ht]
\centering
\caption{Performance of Various CLIP Models on the HRSID Test Set}
\begin{tabular}{lccccccc}
\toprule
Model & i2t-R@1 & i2t-R@5 & i2t-R@10 & t2i-R@1 & t2i-R@5 & t2i-R@10 & Mean Recall \\
\midrule
CLIP           & 0.15 & 0.56 & 1.02  & 0.05 & 0.51 & 0.92  & 0.54 \\
HQRS-CLIP      & 0.00 & 0.36 & 0.61  & 0.05 & 0.46 & 1.07  & 0.42 \\
SAR-CLIP       & 2.09 & 9.64 & 18.71 & 2.29 & 10.91 & 19.58 & 10.54 \\
SAR-RS-CLIP    & \textbf{2.65} & \textbf{10.91} & \textbf{20.50} & \textbf{2.55} & \textbf{11.78} & \textbf{20.75} & \textbf{11.52} \\
\bottomrule
\end{tabular}
\label{tab:hrsid-clip}
\end{table*}

\begin{table*}[ht]
\centering
\caption{Performance of Various CLIP Models on the OSdataset\_512 Test Set}
\begin{tabular}{lccccccc}
\toprule
Model & i2t-R@1 & i2t-R@5 & i2t-R@10 & t2i-R@1 & t2i-R@5 & t2i-R@10 & Mean Recall \\
\midrule
CLIP                & 0.71 & 3.07  & 4.48  & 0.71 & 3.77  & 8.02  & 3.46 \\
HQRS-CLIP           & 1.89 & 5.90  & 8.96  & 3.54 & 8.02  & 12.26 & 6.76 \\
SAR-CLIP            & 4.48 & 14.86 & \textbf{23.82} & 4.72 & \textbf{20.99} & \textbf{29.72} & 16.43 \\
SAR-RS-CLIP         & \textbf{5.66} & \textbf{16.04} & 23.11 & \textbf{5.19} & 20.75 & 28.77 & \textbf{16.59} \\
\bottomrule
\end{tabular}
\label{tab:osdataset-clip}
\end{table*}

This section systematically presents and analyzes the experimental results of the proposed models on caption generation and cross-modal retrieval tasks for SAR imagery. The evaluation covers the transferability and adaptability of SAR-RS-CLIP and SAR-RS-CoCa on SAR images, along with visualization analysis of text-image embeddings. Through both quantitative and qualitative assessments, the effectiveness and generalizability of the SAR-TEXT dataset and the progressive transfer learning strategy are validated in enhancing the multimodal models' semantic understanding and expressive capabilities for SAR scenes.

\subsubsection{SAR-RS-CLIP Cross-Modal Retrieval Experiment}
The section demonstrates the impact of different CLIP training strategies on model performance in SAR-text cross-modal retrieval tasks. Prior to formal training, this paper first evaluates the initial performance of the original CLIP model and the previously fine-tuned HQRS-CLIP model~\cite{he2025enhancingremotesensingvisionlanguage} on SAR images. The experimental results indicate that both the standard CLIP and HQRS-CLIP models exhibit suboptimal cross-modal retrieval performance on SAR images, with an average recall rate of less than 1\% on the test set, thereby validating the limitations of the direct transfer strategy in SAR scenarios.

In the transfer learning experiment, this paper first fine-tunes the CLIP model based on the SAR-TEXT dataset to obtain the SAR-RS-CLIP model. Experimental results show that compared to the original CLIP baseline, SAR-RS-CLIP achieves average recall rate improvements of 10.0\% and 12.97\% on the HRSID and OSdataset\_512 test sets, respectively, significantly validating the effectiveness of the SAR-TEXT dataset in image-text alignment tasks. This result further demonstrates that SAR-TEXT can provide rich and reliable semantic pairing information for SAR images, significantly enhancing cross-modal retrieval performance. Additionally, experiments show that the CLIP model can achieve significant performance improvements in SAR image tasks through simple fine-tuning, confirming the strong transferability of the “pre-training-fine-tuning” paradigm in multi-modal remote sensing tasks.

To further explore more optimal transfer strategies, this paper draws on the empirical analysis in~\cite{13} and proposes a progressive transfer learning scheme. This scheme first pre-trains the CLIP model using remote sensing optical image text data to adapt it to the modal features of the remote sensing field. Subsequently, the model is further fine-tuned to adapt to SAR image tasks, achieving effective knowledge transfer from natural images to SAR images. The model obtained from this two-stage training strategy is named SAR-RS-CLIP. 

As shown in Tables \ref{tab:hrsid-clip} and \ref{tab:osdataset-clip}, The two-stage trained SAR-RS-CLIP model achieved an average recall rate that was 10.98\% and 13.13\% higher than the baseline CLIP model on the HRSID and OSdataset\_512 test sets, respectively, and approximately 1\% and 0.2\% higher than the single-stage SAR-RS-CLIP model. Notably, the performance improvements on the i2t-R@1 and t2i-R@1 metrics of the OSdataset\_512 test set is particularly significant, reaching approximately 1.2\% and 0.5\%, respectively. These results indicate that the progressive transfer learning strategy—pre-training the CLIP model using remote sensing optical image data and then fine-tuning it with SAR-TEXT data—effectively improves the model's cross-modal retrieval performance on SAR images.

\begin{figure}[h]
  \centering
   \includegraphics[width=\linewidth]{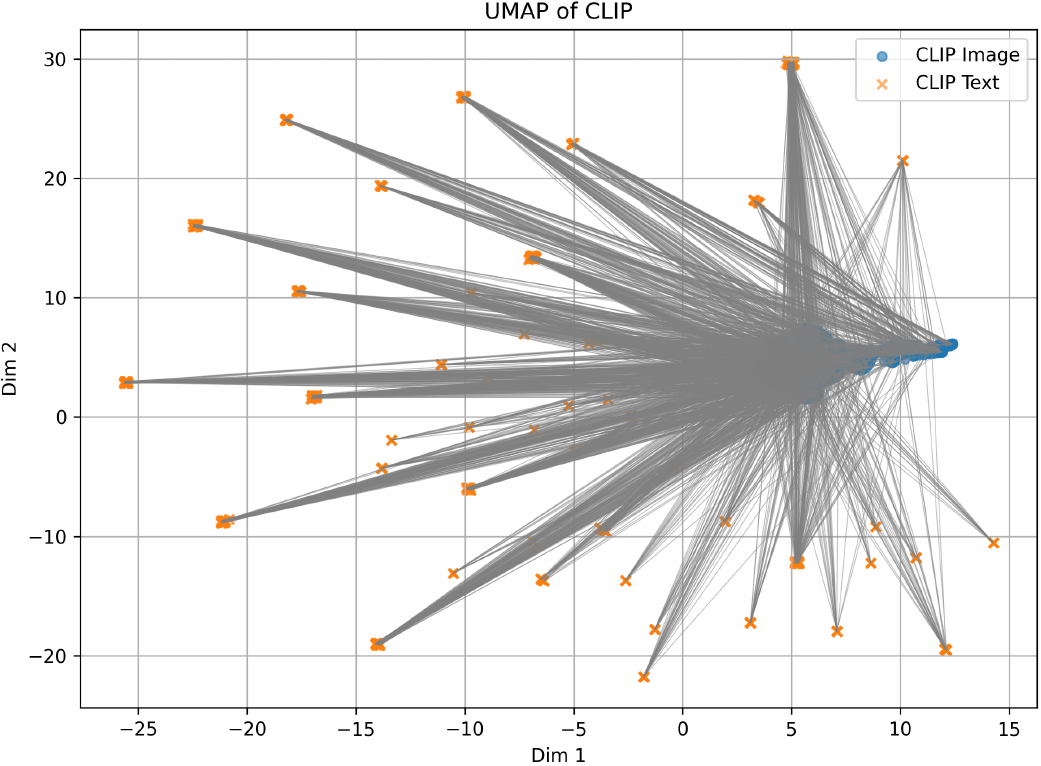}
   \caption{UMAP visualization of text-image embeddings for the HRSID test set using the CLIP model.}
   \label{fig:umap_HRSID_clip}
\end{figure}

\begin{figure}[h]
  \centering
   \includegraphics[width=\linewidth]{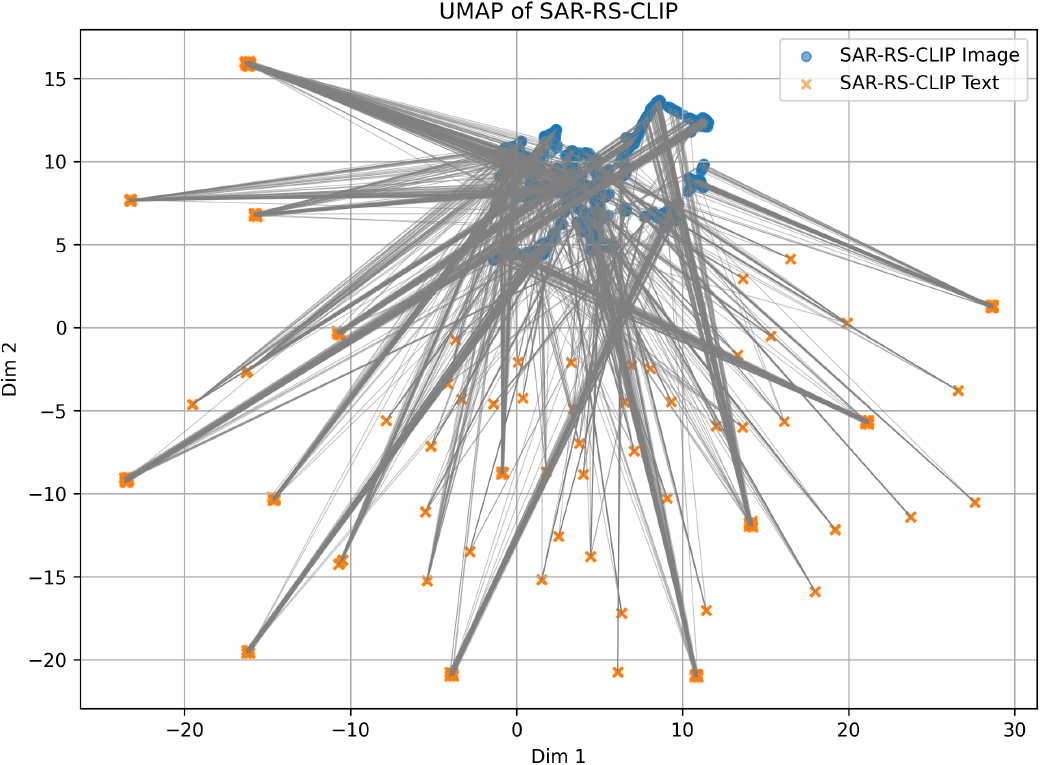}
   \caption{UMAP visualization of text-image embeddings for the HRSID test set using the SAR-RS-CLIP model.}
   \label{fig:umap_HRSID_sar_rs_clip}
\end{figure}

\subsubsection{Visualization of image and text embedding based on UMAP}
To intuitively evaluate the performance of CLIP and SAR-RS-CLIP models in SAR image-text representation and cross-modal alignment tasks, this paper uses the UMAP algorithm to reduce the dimensionality of image and text embeddings in the HRSID and OSdataset\_512 test sets for visualization, mapping the distribution in the high-dimensional semantic space onto a two-dimensional plane to facilitate observation of the structural differences and alignment capabilities of different models in the embedding space.

Figures \ref{fig:umap_HRSID_clip} and \ref{fig:umap_HRSID_sar_rs_clip} present the UMAP visualization results of the CLIP and SAR-RS-CLIP models on the HRSID test set. Due to the high homogeneity of the HRSID dataset, all images feature ship targets with similar appearance and texture characteristics, which makes it challenging for the model to effectively distinguish between them. As shown in the figure, the image distribution of the CLIP model is more concentrated, suggesting that the images are challenging for the model to distinguish. After training, the image distribution of the two-stage SAR-RS-CLIP model becomes significantly more dispersed, indicating that the model's ability to differentiate between images in the HRSID dataset has improved to some extent. Nevertheless, despite the limited overall separability, the SAR-RS-CLIP model demonstrates clearer image distribution boundaries compared to CLIP and achieves better separation among image embeddings. This indicates that its capacity to capture fine-grained differences is enhanced to some extent through SAR-TEXT training.

\begin{figure}[h]
  \centering
   \includegraphics[width=\linewidth]{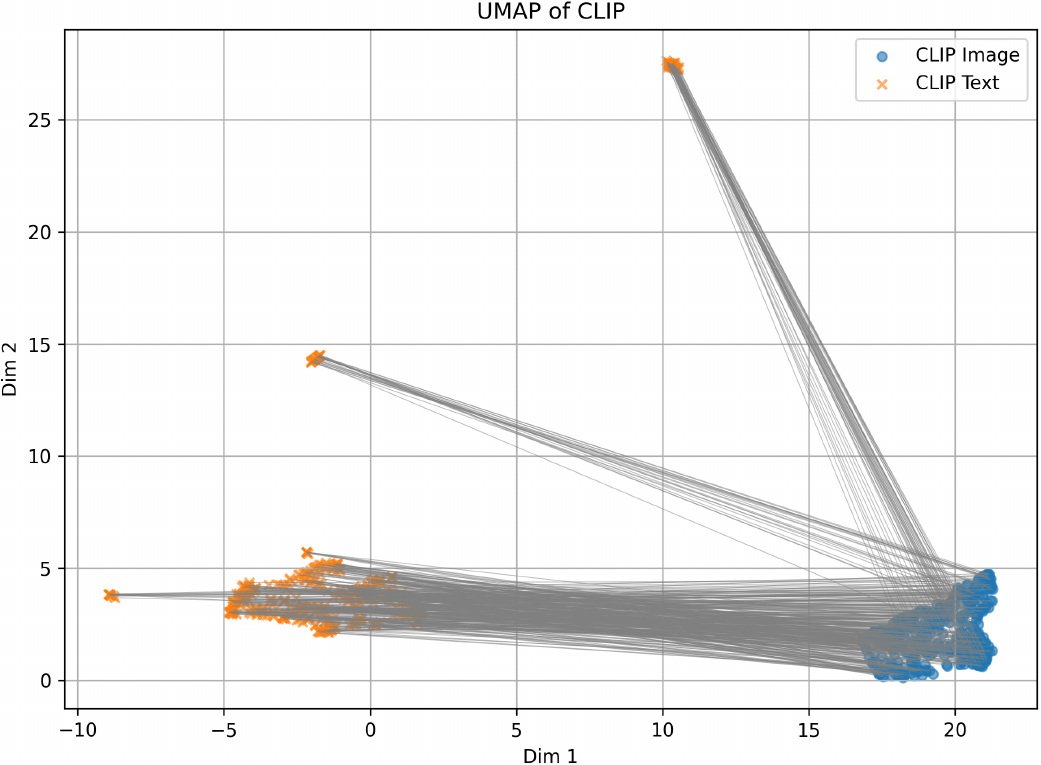}
   \caption{UMAP visualization of text-image embeddings for the OSdataset\_512 test set using the CLIP model.}
   \label{fig:umap_os_clip}
\end{figure}

\begin{figure}[h]
  \centering
   \includegraphics[width=\linewidth]{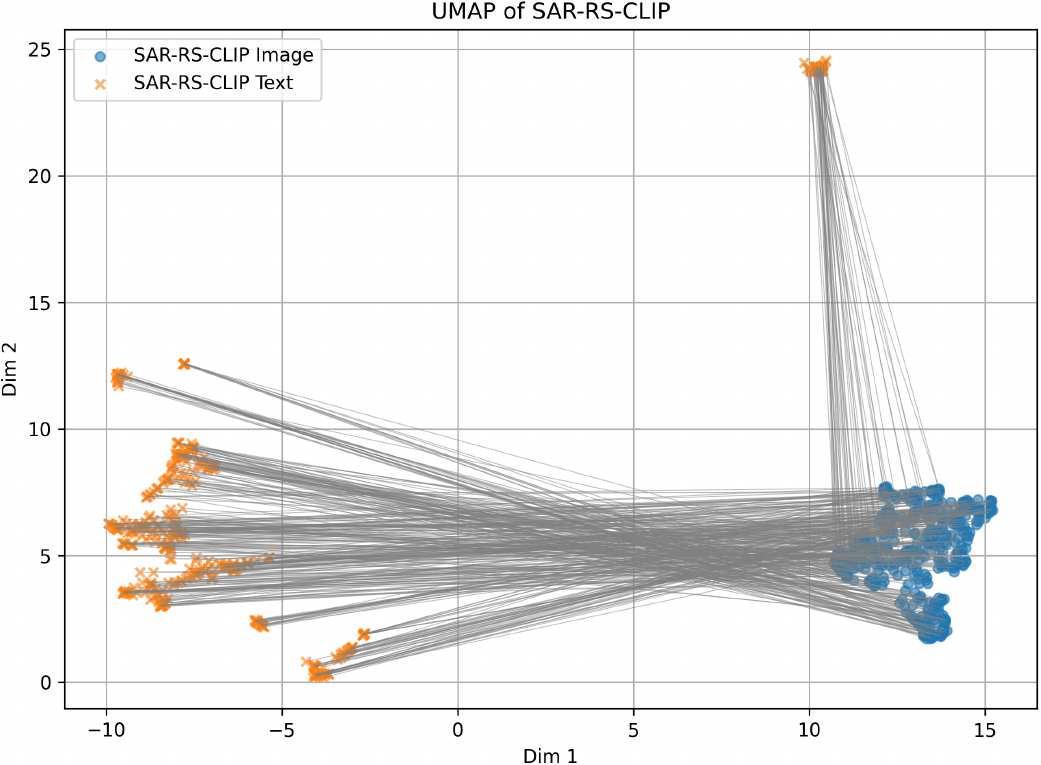}
   \caption{UMAP visualization of text-image embeddings for the OSdataset\_512 test set using the SAR-RS-CLIP model.}
   \label{fig:umap_os_sar_rs_clip}
\end{figure}

Figures \ref{fig:umap_os_clip} and \ref{fig:umap_os_sar_rs_clip} display the text-image embedding visualizations of the two models on the OSdataset\_512 test set. Compared to HRSID, the image samples in OSdataset\_512 exhibit greater diversity, resulting in a more dispersed and structurally rich distribution of embeddings in the two-dimensional space. Notably, the SAR-RS-CLIP model significantly improves the distinguishability between images and text on this dataset, with embedding clusters of different categories showing more distinct boundaries and clearer text-image pairing structures. This demonstrates its enhanced capabilities in cross-modal alignment and semantic representation, which can contribute to improved performance in downstream text-image retrieval tasks.

\begin{table*}[ht]
\centering
\caption{Evaluation Results of CoCa Models on the HRSID Test Set}
\begin{tabular}{lcccccccc}
\toprule
Model & SPICE & BLEU-1 & BLEU-2 & BLEU-3 & BLEU-4 & METEOR & ROUGE-L & CIDEr \\
\midrule
CoCa         & 0.043 & 0.092 & 0.036 & 0.009 & 0.003 & 0.077 & 0.093 & 0.005 \\
RS-CoCa      & 0.059 & 0.081 & 0.039 & 0.014 & 0.007 & 0.088 & 0.121 & 0.001 \\
SAR-CoCa     & \textbf{0.689} & 0.680 & 0.624 & 0.571 & 0.519 & 0.522 & 0.781 & 2.513 \\
SAR-RS-CoCa  & 0.688 & \textbf{0.694} & \textbf{0.637} & \textbf{0.583} & \textbf{0.530} & \textbf{0.523} & \textbf{0.792} & \textbf{3.186} \\
\bottomrule
\end{tabular}
\label{tab:hrsid-coca}
\end{table*}

\begin{table*}[ht]
\centering
\caption{Evaluation Results of CoCa Models on the OSdataset\_512 Test Set}
\begin{tabular}{lcccccccc}
\toprule
Model & SPICE & BLEU-1 & BLEU-2 & BLEU-3 & BLEU-4 & METEOR & ROUGE-L & CIDEr \\
\midrule
CoCa         & 0.022 & 0.118 & 0.026 & 0.004 & 0.000 & 0.040 & 0.109 & 0.019 \\
RS-CoCa      & 0.054 & 0.095 & 0.187 & 0.091 & 0.053 & 0.035 & 0.132 & 0.205 \\
SAR-CoCa     & 0.282 & 0.409 & 0.301 & 0.243 & 0.198 & 0.211 & 0.391 & 0.631 \\
SAR-RS-CoCa  & \textbf{0.298} & \textbf{0.420} & \textbf{0.319} & \textbf{0.263} & \textbf{0.219} & \textbf{0.228} & \textbf{0.410} & \textbf{0.665} \\
\bottomrule
\end{tabular}
\label{tab:os-coca}
\end{table*}

\subsubsection{SAR-RS-CoCa Image Caption Generation Test Experiment}
Although the CLIP model has achieved significant performance improvements in SAR image recognition tasks, it is essentially a discriminative model and cannot directly generate text from images. To further expand the automated interpretation capabilities of SAR images, this paper explores its application in generative vision-language models based on the SAR-TEXT dataset, with the aim of training a model that can directly convert SAR images into natural language descriptions.

Given the CoCa model's outstanding performance in image caption generation and cross-modal tasks, as well as its structurally simple advantages, this paper selects CoCa as the fine-tuning foundation for the generative model. Similar to CLIP, CoCa also employs contrastive learning to align image and text features. However, its key distinction lies in the introduction of a generative decoder module, combined with language modeling loss (generative loss), further enhancing the semantic alignment capability between images and text.

Table \ref{tab:hrsid-coca} and Table \ref{tab:os-coca} present the performance of different CoCa models on the HRSID and OSdataset test sets. First, this paper evaluates the zero-shot capability of the pre-trained CoCa model and the RS-CoCa model trained by He et al.~\cite{he2025enhancingremotesensingvisionlanguage} in the task of caption generation for SAR images. The experimental results indicate that the CoCa model, which has not been trained specifically for SAR images, performs extremely poorly in this task, barely able to generate effective captions. This suggests that without the ability to align SAR image and text semantics, the model struggles to generalize to the SAR image domain.

Based on this, this paper draws on the CLIP fine-tuning paradigm and first uses the SAR-TEXT dataset to fine-tune the CoCa model, resulting in the SAR-CoCa model. Furthermore, this paper introduces a progressive transfer learning strategy and adopts a two-stage approach of “first pre-training based on optical remote sensing images, then fine-tuning based on the SAR-TEXT dataset” to train the SAR-RS-CoCa model.

The experimental results show that the performance of the two generative models fine-tuned with SAR text-image data has significantly improved on the HRSID and OSdataset test sets, with performance enhancements in multiple metrics reaching up to several times, fully validating the effectiveness and value of the SAR-TEXT dataset in generative tasks. Additionally, SAR-RS-CoCa outperformed SAR-CoCa on both test sets, further confirming the adaptability and advantages of the two-stage transfer learning strategy in generative tasks.

\subsubsection{SAR-RS-CoCa Caption Generation Qualitative Assessment}
\begin{figure}[t]
  \centering
   \includegraphics[width=\linewidth]{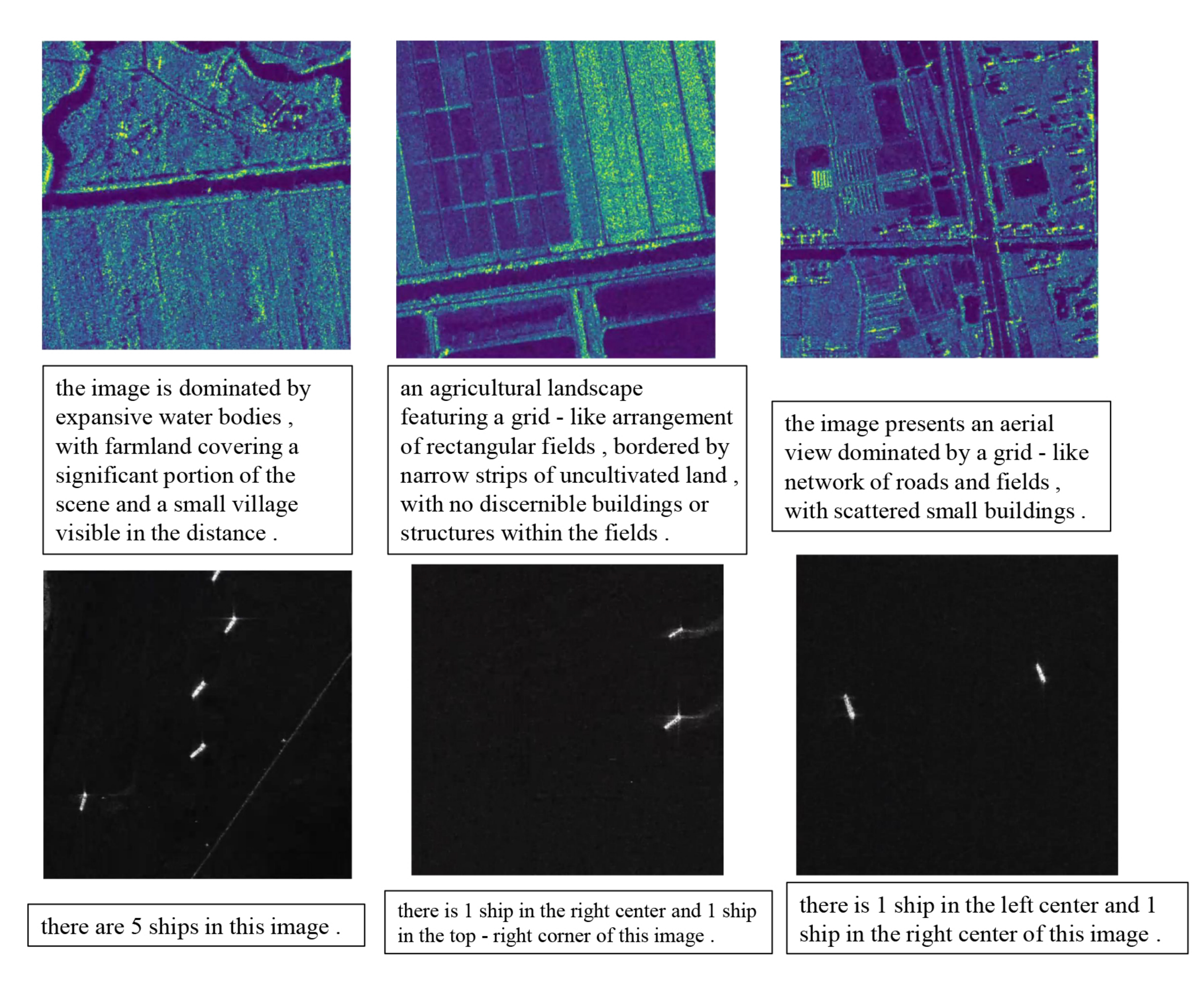}
   \caption{SAR-RS-CoCa Caption Examples Generated for Various SAR Images.}
   \label{fig:sar-rs-coca}
\end{figure}

To more intuitively demonstrate the performance of the SAR-RS-CoCa model in the image caption generation task, this paper selected several representative SAR image samples for qualitative analysis. As shown in Figure \ref{fig:sar-rs-coca}, the SAR-RS-CoCa model demonstrates excellent generation capabilities in various types of SAR image scenarios, producing English description sentences with reasonable structure and accurate semantics.

\begin{table*}[ht]
\centering
\caption{Comparison of Two Training Strategies on the HRSID Test Set}
\begin{tabular}{lccccccc}
\toprule
Model & i2t-R@1 & i2t-R@5 & i2t-R@10 & t2i-R@1 & t2i-R@5 & t2i-R@10 & Mean Recall \\
\midrule
Single-SAR-RS-CLIP & 2.24 & 8.92  & 16.57 & 1.99 & 9.74  & 17.19 & 9.44 \\
SAR-RS-CLIP        & \textbf{2.65} & \textbf{10.91} & \textbf{20.50} & \textbf{2.55} & \textbf{11.78} & \textbf{20.75} & \textbf{11.52} \\
\bottomrule
\end{tabular}
\label{tab:hrscd-comparison}
\end{table*}

\begin{table*}[ht]
\centering
\caption{Comparison of Two Training Strategies on the OSdataset\_512 Test Set}
\begin{tabular}{lccccccc}
\toprule
Model & i2t-R@1 & i2t-R@5 & i2t-R@10 & t2i-R@1 & t2i-R@5 & t2i-R@10 & Mean Recall \\
\midrule
Single-SAR-RS-CLIP & 2.83 & 11.32 & 19.34 & 4.01 & 15.80 & 25.71 & 13.17 \\
SAR-RS-CLIP        & \textbf{5.66} & \textbf{16.04} & \textbf{23.11} & \textbf{5.19} & \textbf{20.75} & \textbf{28.77} & \textbf{16.59} \\
\bottomrule
\end{tabular}
\label{tab:osdataset-comparison}
\end{table*}

In natural scene images (top row), the model successfully extracts key semantic elements from the images, such as water bodies, farmland, road networks, and villages, and generates complete sentences with strong linguistic organization and semantic hierarchy. This demonstrates that SAR-RS-CoCa not only has the ability to perceive low-level texture features but also has the ability to map these features to high-level semantic information, fully demonstrating its potential in remote sensing natural scene understanding.

In target detection-type images (bottom row), despite the lack of traditional visual features such as color and texture in SAR images, and the relatively simple scene background, the model is still able to accurately identify ships in the images and reasonably locate their relative positions. For example, the model can accurately output detailed descriptions such as “There are 5 ships in the image” and “There is 1 ship in the middle of the right side of the image, and another 1 in the top-right corner,” demonstrating its strong capabilities in estimating the number of objects and understanding their spatial orientation.

However, it is important to note that SAR-RS-CoCa's subtitle generation capabilities primarily stem from its learning of training data. When the model is trained solely on the SAR-TEXT dataset, the generated subtitles often exhibit a noticeable “style bias,” meaning the generated results closely resemble the training corpus in terms of language expression style and descriptive methods. This stylistic consistency to some extent limits the model's adaptability in diverse application scenarios. Therefore, to enhance SAR-Captioner's generalization capability in subtitle generation tasks, future research should focus on constructing training datasets with more diverse styles and exploring generalization mechanisms that can enhance the model's language style transfer and scenario adaptability.

In summary, SAR-RS-CoCa demonstrates strong capabilities in SAR image captioning. For natural scene imagery, it can generate coherent descriptions that capture key global semantic elements such as land cover types and structural patterns. For images containing specific targets, the model can produce textual descriptions that include not only object categories but also indicative positional or quantitative information, thus enriching the interpretability of the captions. These results highlight the effectiveness of the SAR-TEXT dataset in supporting caption generation for SAR images, and also suggest that the progressive transfer learning strategy provides a promising approach for enhancing the semantic expressiveness of generative models in SAR-related applications.

\subsection{Ablation Study}

\begin{table*}[]
\centering
\caption{Experimental results of SAR-GPT on the HRSID test set}
\begin{tabular}{lcccccccc}
\toprule
Model & SPICE & BLEU-1 & BLEU-2 & BLEU-3 & BLEU-4 & METEOR & ROUGE-L & CIDEr \\
\midrule
TinyGPT-V & 0.1789 & 0.1969 & 0.0844 & 0.0391 & 0.0000 & 0.1296 & 0.1975 & 0.2146 \\
S234-SAR  & 0.2959 & 0.3240 & 0.2295 & 0.1740 & 0.1328 & 0.2239 & 0.2975 & 0.9667 \\
SAR-GPT   & \textbf{0.3211} & \textbf{0.3450} & \textbf{0.2603} & \textbf{0.2086} & \textbf{0.1716} & \textbf{0.2350} & \textbf{0.3341} & \textbf{1.4732} \\
\bottomrule
\end{tabular}
\label{tab:sar_gpt_hrsid}
\end{table*}

\begin{table*}[]
\centering
\caption{Experimental results of SAR-GPT on the OSdataset\_256 test set}
\begin{tabular}{lcccccccc}
\toprule
Model & SPICE & BLEU-1 & BLEU-2 & BLEU-3 & BLEU-4 & METEOR & ROUGE-L & CIDEr \\
\midrule
TinyGPT-V & 0.1269 & 0.1487 & 0.0732 & 0.0405 & 0.0215 & 0.1093 & 0.1332 & 0.0483 \\
S234-SAR  & 0.1669 & 0.2547 & 0.1372 & 0.0943 & 0.0730 & 0.1554 & 0.1942 & 0.5513 \\
SAR-GPT   & \textbf{0.2974} & \textbf{0.3347} & \textbf{0.2626} & \textbf{0.2285} & \textbf{0.2057} & \textbf{0.1999} & \textbf{0.3276} & \textbf{1.7976} \\
\bottomrule
\end{tabular}
\label{tab:sar_gpt_osdataset256}
\end{table*}

To evaluate the effectiveness of the proposed progressive fine-tuning strategy in enhancing CLIP’s performance on cross-modal retrieval tasks involving SAR imagery, a comparative experiment is conducted. Two training strategies are compared: single-stage mixed training (Single-SAR-RS-CLIP) and two-stage progressive training (SAR-RS-CLIP). Experiments are performed on two test datasets, HRSID and OSdataset\_512, using multiple recall-based metrics—including i2t-R@K, t2i-R@K, and mean recall as evaluation criteria.

\begin{itemize}
    \item \textbf{Single-stage mixed training (Single-SAR-RS-CLIP):} This training scheme integrates 210,000 optical remote sensing image–text pairs from the HQRS-IT-210K~\cite{he2025enhancingremotesensingvisionlanguage} dataset and 130,000 SAR image–text pairs from the SAR-TEXT dataset to jointly train the RS-CLIP model. All samples are uniformly fed into the model, with the data loader randomly selecting instances from the combined dataset in each training batch. The resulting model is referred to as Single-SAR-RS-CLIP.
    
    \item \textbf{Two-stage progressive training (SAR-RS-CLIP):}This training scheme first continual pretraining the CLIP model on the HQRS-IT-210K dataset to adapt it to the optical remote sensing image–text matching task. The model is then further fine-tuned using the SAR-TEXT dataset to complete the transfer to SAR image–text matching. The resulting model is referred to as SAR-RS-CLIP.
\end{itemize}

Tables \ref{tab:hrscd-comparison} and \ref{tab:osdataset-comparison} show the performance comparison of the two training strategies on the HRSID and OSdataset\_512 test sets. The evaluation metrics cover the recall rates R@1, R@5, and R@10 for both image retrieval to text (i2t) and text retrieval to image (t2i), as well as the overall mean recall.

On the HRSID test set, SAR-RS-CLIP outperforms Single-SAR-RS-CLIP across all metrics. In the i2t direction, all three recall metrics show improvements, with a relative increase of +3.93\% in i2t-R@10. Improvements in the t2i direction are equally notable, with the t2i-R@10 metric improving by +3.56\%. The overall average recall rate increased from 9.44\% to 11.52\%, representing a relative increase of 22.0\%, indicating that the two-stage training strategy demonstrates superior semantic alignment and retrieval capabilities in medium-to-low resolution remote sensing scenarios.

On the OSdataset\_512 dataset, SAR-RS-CLIP also achieves comprehensive leadership across all metrics, particularly demonstrating stronger retrieval performance in high-precision metrics such as i2t-R@1 (+2.83\%) and t2i-R@10 (+3.06\%). The mean recall improves from 13.17\% to 16.59\%, with an absolute increase of +3.42\% and a relative increase of approximately 26.0\%, further validating the generalization capability of the incremental strategy in high-resolution, complex scenarios.

In summary, SAR-RS-CLIP outperforms single-stage training models on all core evaluation metrics across two test datasets, significantly improving cross-modal retrieval performance. Experimental results indicate that by first adapting the model to remote sensing optical image pairs and then transferring it to SAR image tasks, modal differences can be effectively mitigated, enhancing the model's robustness and generalization capabilities under strongly heterogeneous and low-resource conditions, thereby further leveraging the cross-modal representation potential of the CLIP model.

\subsection{Further exploration}

To further explore the practicality of the proposed SAR-TEXT dataset, we construct SAR-VQA, an instruction-based dataset for SAR images derived from SAR-TEXT. SAR-VQA consists of over 130,000 simple instruction pairs and approximately 450,000 conversational instruction samples, designed to enhance the multimodal understanding capabilities of MLLMs for SAR image interpretation. Based on the SAR-VQA dataset, we introduce the TinyGPT-V model~\cite{yuan2023tinygpt} for fine-tuning and train SAR-GPT for SAR image question-answering tasks. The performance of SAR-GPT is evaluated from both quantitative and qualitative perspectives.

\subsubsection{Quantitative Evaluation of SAR-GPT}
To evaluate the capability of SAR-GPT in the SAR image visual question answering (VQA) task, we randomly select 100 image–text pairs from the test sets of the HRSID and OSdataset\_256 datasets, respectively. These samples are processed following the SAR-VQA construction protocol, resulting in two SAR-VQA test subsets for model evaluation. We assess the performance of three multimodal large language models (MLLMs) on these test sets: TinyGPT-V, S234-SAR, and SAR-GPT. Specifically, TinyGPT-V refers to the original model without any fine-tuning; S234-SAR is trained using a single-stage approach with fine-tuning conducted solely on the SAR-VQA dataset; and SAR-GPT represents the fully fine-tuned TinyGPT-V model obtained through a two-stage training process.

Tables \ref{tab:sar_gpt_hrsid} and \ref{tab:sar_gpt_osdataset256} present the performance of each model on the two test sets, respectively. A comparison between the experimental results of SAR-GPT and S234-SAR demonstrates that the two-stage training strategy provides a clear advantage in SAR image generation tasks. SAR-GPT consistently outperforms S234-SAR on both the HRSID and OSdataset\_256 test sets, with particularly notable improvements in generative evaluation metrics such as CIDEr. These results validate the effectiveness of the progressive fine-tuning approach in cross-modal tasks within the SAR domain, successfully enhancing the alignment between SAR images and large language models (LLMs) such as TinyGPT-V.

\begin{figure}[t]
  \centering
   \includegraphics[width=0.75\linewidth]{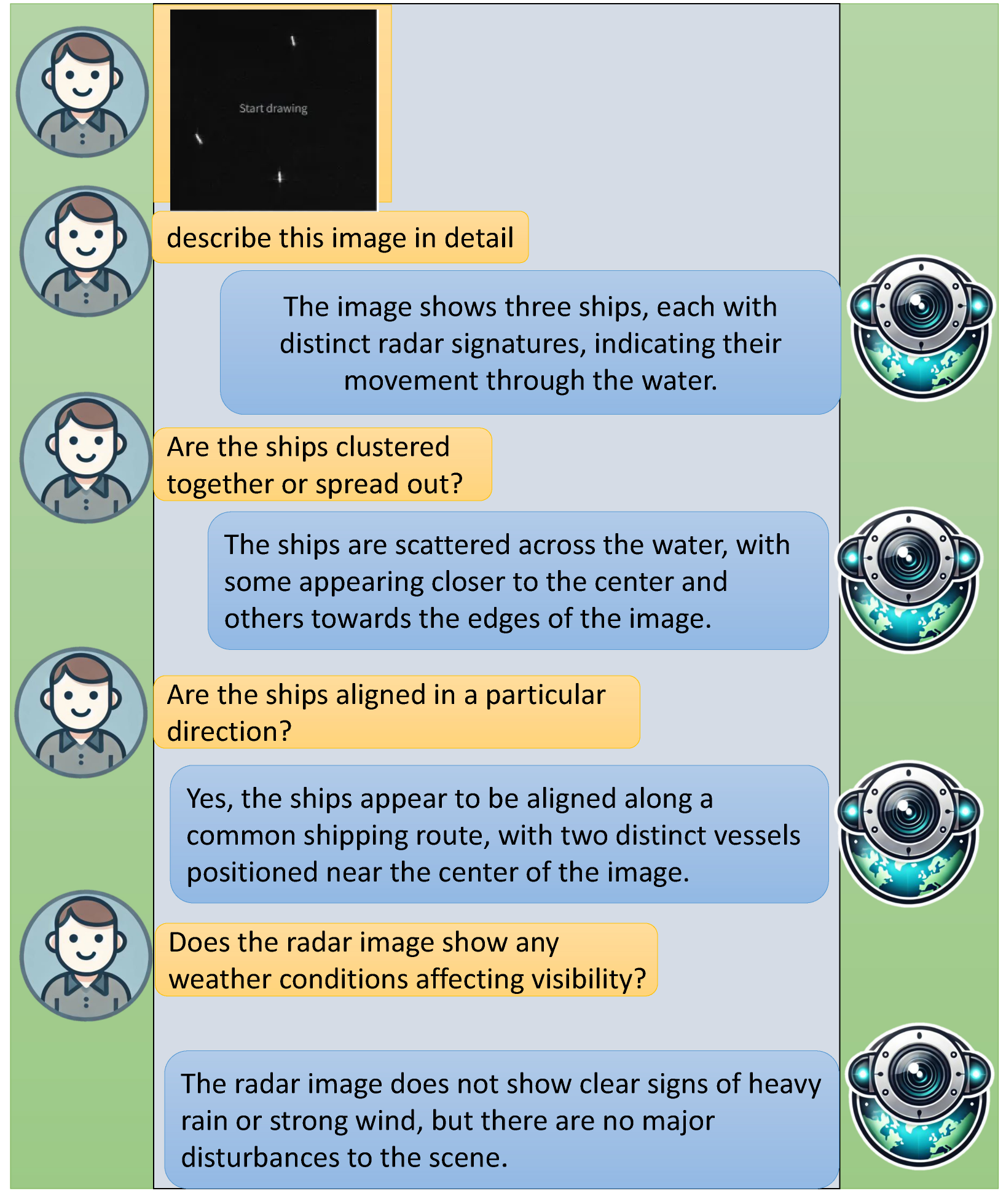}
   \caption{Example of Human–Machine Dialogue with SAR-GPT.}
   \label{fig:sar_gpt_demo}
\end{figure}

\subsubsection{Qualitative Analysis of SAR-GPT of SAR-GPT}
Given the current lack of authoritative quantitative evaluation datasets, intuitive qualitative assessments become particularly important. Figure \ref{fig:sar_gpt_demo} shows an example of SAR-GPT performing multi-turn SAR image–text question answering. The dialogue demonstrates the model’s ability to understand SAR image details, respond to complex queries, and generate semantically consistent natural language descriptions. These results highlight SAR-GPT’s effective cross-modal reasoning and alignment between SAR images and language.

Overall, our preliminary quantitative and qualitative observations suggest that SAR-GPT can leverage the SAR-TEXT dataset to produce reasonable image–text responses, indicating the potential of SAR-TEXT to support GPT-style multimodal models in the SAR domain. However, the current results should be interpreted as exploratory rather than conclusive: The SAR-GPT training process remains relatively simple, the evaluation benchmarks are limited in scale and coverage, and the generated answers sometimes show stylistic bias or omit fine-grained details. We regard this study as an initial attempt to probe the practicality of SAR-TEXT for large language model adaptation, while leaving a more systematic exploration of SAR-GPT training and evaluation to future work.

\section{Conclusion}

Given the current lack of large-scale SAR image–text paired datasets, this study systematically introduces SAR-TEXT, a large-scale and high-quality dataset designed to address the challenge of semantic interpretation of SAR images. To support dataset construction, we propose SAR-Narrator, a caption auto-generation framework that transforms structured labels into high-quality natural language descriptions. Building upon SAR-TEXT, we further develop a progressive fine-tuning strategy that first adapts models on optical remote sensing data, followed by fine-tuning on SAR image–text pairs. This strategy is employed to train two vision–language foundation models: SAR-RS-CLIP for cross-modal retrieval and SAR-RS-CoCa for caption generation. Experimental results show that the proposed framework significantly outperforms existing methods across multiple benchmarks, validating both the effectiveness of SAR-Narrator and the feasibility of the proposed transfer learning strategy. Moreover, through exploration of the SAR-VQA task, we demonstrate that SAR-TEXT not only serves as a valuable resource for automatic semantic interpretation of SAR images, but also lays a solid foundation for advancing multimodal intelligent interpretation in the remote sensing field.

{
    \small
    \bibliographystyle{ieeenat_fullname}

    \bibliography{main}
}

\end{document}